%% file: main.tex
\documentclass[10pt,twocolumn,letterpaper]{article}

\usepackage{cvpr}              

\input{preamble}

\definecolor{cvprblue}{rgb}{0.21,0.49,0.74}
\usepackage[pagebackref,breaklinks,colorlinks,citecolor=cvprblue]{hyperref}

\def\paperID{5312} 
\def\confName{CVPR}
\def\confYear{2024}

\title{From Summary to Action: Enhancing Large Language Models for Complex Tasks with Open World APIs}

\usepackage[ruled,linesnumbered]{algorithm2e}
\usepackage{algorithmic}

\def\ie{\textit{i.e.}}

\def\etc{\textit{etc.}}

\author{{Yulong Liu\textsuperscript{\rm 1}\thanks{Equal contribution}}\quad
Yunlong Yuan\textsuperscript{\rm 2}$^{\ast}$\quad
Chunwei Wang\textsuperscript{\rm 3}\quad
Jianhua Han\textsuperscript{\rm 3}\quad
Yongqiang Ma\textsuperscript{\rm 1}\\
Li Zhang\textsuperscript{\rm 2}\quad
Nanning Zheng\textsuperscript{\rm 1}\thanks{Corresponding author}\quad
Hang Xu\textsuperscript{\rm 3}
\vspace{0.4em}
\\
{\small 
\textsuperscript{\rm 1}National Key Laboratory of Human-Machine Hybrid Augmented Intelligence, Xi'an, China
}\\
{\small
\textsuperscript{\rm 2} School of Data Science, Fudan University
}\quad 
{\small
\textsuperscript{\rm 3} Huawei Noah’s Ark Lab
}
}

\begin{document}
\maketitle
\input{sec/0_abstract}    
\input{sec/1_intro}

\input{sec/2_related}
\input{sec/3_method_new}

\input{sec/4_exp}

\input{sec/5_conclusion}

{
    \small
    \bibliographystyle{ieeenat_fullname}
    \bibliography{main}
}

\input{sec/X_suppl}

\end{document}

%% file: preamble.tex
\usepackage[dvipsnames]{xcolor}

\usepackage{multirow}

%% file: sec/0_abstract.tex
\begin{abstract}
The distinction between humans and animals lies in the unique ability of humans to use and create tools.
Tools empower humans to overcome physiological limitations, fostering the creation of magnificent civilizations.
Similarly, enabling foundational models like Large Language Models (LLMs) with the capacity to learn external tool usage may serve as a pivotal step toward realizing artificial general intelligence.
Previous studies in this field have predominantly pursued two distinct approaches to augment the tool invocation capabilities of LLMs. The first approach emphasizes the construction of relevant datasets for model fine-tuning. The second approach, in contrast, aims to fully exploit the inherent reasoning abilities of LLMs through in-context learning strategies.
In this work, we introduce a novel tool invocation pipeline designed to control massive real-world APIs. This pipeline mirrors the human task-solving process, addressing complicated real-life user queries.
At each step, we guide LLMs to summarize the achieved results and determine the next course of action.
We term this pipeline `from Summary to action', {\bf Sum2Act} for short.
Empirical evaluations of our Sum2Act pipeline on the ToolBench benchmark show significant performance improvements, outperforming established methods like ReAct and DFSDT. This highlights Sum2Act's effectiveness in enhancing LLMs for complex real-world tasks.

\end{abstract}

%% file: sec/1_intro.tex
\section{Introduction}
\label{sec:intro}
\begin{figure}[t]
\begin{center}
   \includegraphics[width=1.0\linewidth]{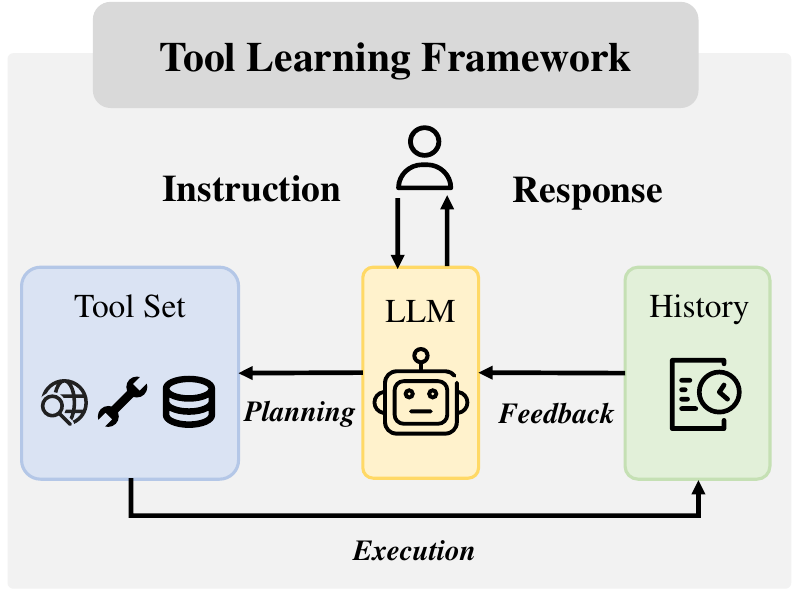}
\end{center}
   \caption{Overview of the general tool learning framework, which consists of three components: LLM, tool set, and history. The user provides an instruction to the LLM, which iteratively selects tools from the tool set, and records results in the history until it deems the instruction complete. The LLM then responds to the user.}
\label{fig:intro}
\end{figure}

Humans possess an exceptional ability to create and utilize tools, transcending their physical constraints and venturing into new realms. 
The recent advancements in Large Language Models (LLMs) such as T5~\cite{T5_2020t5}, LLaMA~\cite{LLaMA_touvron2023llama}, ChatGPT~\cite{2022chatgpt}, GPT4~\cite{2023gpt4}, and others, have showcased their remarkable abilities in comprehending human language.

The emergence of these powerful LLMs marks a significant step towards enabling artificial intelligence systems to match human proficiency in tool use. 
While tasks like computations and drawing pictures might pose a challenge to standalone LLMs, they become feasible when paired with specialized tools. By integrating with a variety of tools, LLMs can interpret and process data not just in text form, but across multiple modalities including visual and auditory modality. 
Visual ChatGPT~\cite{VisualChatGPT_wu2023visual} has successfully pioneered the integration of a large Language Model with visual foundation models~\cite{blip_li2022blip,StableDiffusion_rombach2022high,controlnet_zhang2023adding,instructpix2pix_brooks2023instructpix2pix,sam_kirillov2023segany,gDINO_liu2023grounding} for task completion.
HuggingGPT~\cite{HuggingGPT_shen2023hugginggpt} goes beyond by directly accessing models from the Hugging Face website and ControLLM~\cite{ControlLLM_2023controlllm} expands its capabilities to include audio perception and pointing inputs compared to HuggingGPT.
The capability of using tools is crucial in the development of artificial general intelligence. 
Consequently, the augmentation of LLMs with diverse tools has become an increasingly focal area of research in recent times~\cite{ToolLearning_qin2023tool}.

Figure~\ref{fig:intro} presents an overview of Tool Learning, highlighting its core principle: understanding the functionality of various tools and employing them effectively to accomplish specific tasks.
Prior studies in Tool Learning have been constrained to a limited and static set of tools, often neglecting the integration of real-world APIs or considering only a narrow range of APIs with limited diversity~\cite{Gorilla_patil2023gorilla, ToolAlpaca_tang2023toolalpaca, APIbank_li2023api}. 
These approaches typically employed either the ReACT~\cite{ReACT_yao2022react} or Chain of Thought~\cite{ChainOfThought_wei2022chain} methods for reasoning. However, such methods do not fully exploit the capabilities of Large Language Models, leading to challenges in addressing complex tasks.
In the case of ToolLLM~\cite{toolllama_qin2023toolllm}, the DFSDT (Depth First Search-based Decision Tree) approach was introduced to tackle complex real-world tasks, accompanied by a pioneering benchmark called ToolBench, 
which contains over 16,000 real-world APIs across 49 categories. Nevertheless, we observed that DFSDT tends to overlook valuable information from failed paths while exploring new branches.
In this work, we focus on the effective use of massive open-world APIs for LLMs. In this setting, the responses of APIs are dynamic, which means that they do not always give the expected information and the LLMs should have the ability to deal with failures and find new paths.

Our method is inspired by the strategies people use to accomplish tasks. 
Typically, when an individual is presented with an instruction, they select appropriate tools to accomplish the related task. As each subtask is completed, the person notes the information gained and progresses to the next part of the task, continuing in this manner until the entire task is completed.
Crucially, if a subtask proves unsolvable with the initially chosen tool, the individual will attempt alternative tools or, if necessary, decide to give up. 
In essence, a rational human always maintains a clear understanding of their objective, the current state of the task, and the subsequent steps to take, and importantly, recognizes when it is prudent to terminate the process.


In prior research, the task state has been implicitly indicated through past paths~\cite{ChainOfThought_wei2022chain, toolllama_qin2023toolllm}. 
However, LLMs often struggle to summarize essential information in lengthy contexts. To address this, we propose a method where the LLM summarizes historical information at each step before deciding the next action. This summarization process offers two key advantages: (1) it keeps the context length within a manageable range, and (2) it helps the LLM explicitly understand its current task state. We have named this approach {\bf Sum2Act}.

Upon receiving user instructions, Sum2Act operates through two main components: the router and the state manager.
The router, after considering the target task and the current task state (which includes both progress and failure information), decides on the next step to be taken. 
Simultaneously, the state manager updates the task state based on the outcomes of new actions.
Additionally, we have integrated a reflection capability into the state manager, enabling the LLM to identify and correct errors that arise during the task-solving process.

Our pipeline is evaluated on ToolBench~\cite{toolllama_qin2023toolllm}, a benchmark that aggregates over 16,000 real-world APIs across 49 categories.
Results from these tests indicate that under our framework, the model demonstrates enhanced performance, effectively handling a wide range of tasks.

In general, Our contributions are as follows:
\begin{itemize}
\item We introduce a novel tool invocation framework comprising a router and a state manager. This framework equips Large Language Models (LLMs) with the capability to explicitly monitor task progress and rectify errors.
\item Comparative experiments on ToolBench reveal that our framework outperforms existing baselines, including CoT and DFSDT, demonstrating the effectiveness of our method in addressing complicated real-world tasks.
\item Visual APIs additionally are incorporated with open-world APIs in ToolBench, showing the ability of Sum2Act to handle more diverse vision tasks.
\end{itemize}

%% file: sec/2_related.tex
\begin{figure*}[t]
\begin{center}
   \includegraphics[width=1.0\linewidth]{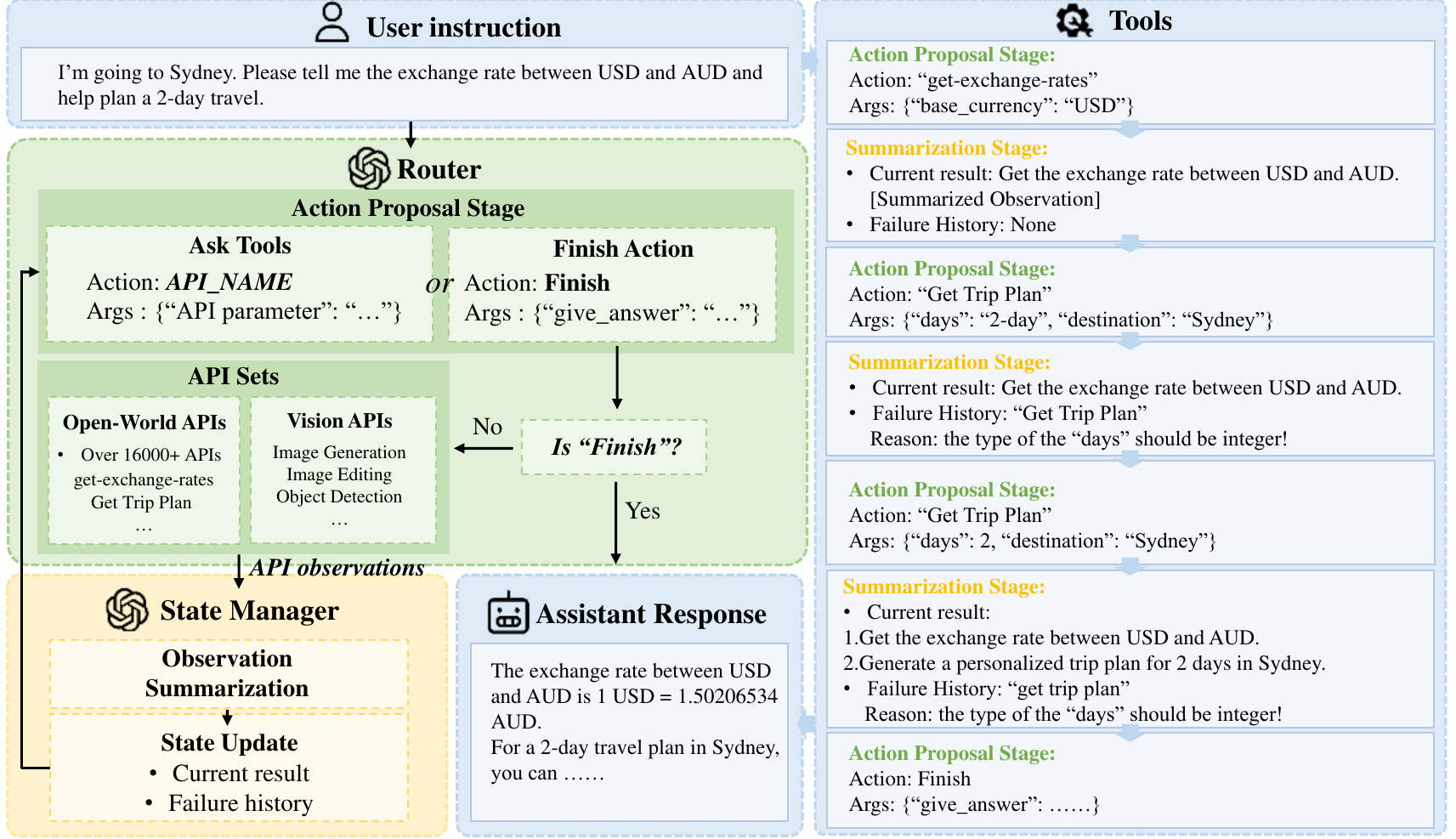}
\end{center}
   \caption{The Framework of Sum2Act: Components and Workflow. The router decides the next action based on user instruction and the current state, then finds and runs the right API. The State Manager summarizes the observation from APIs and updates the state with information from the API. The process ends when the router decides the goal is reached and provides the answer to the user.}
\label{fig:pipeline}
\end{figure*}

\section{Related Work}
\label{sec:related_work}

\paragraph{Tool Learning.}
Recent studies reveal the promising capabilities of LLMs to manipulate tools, \ie, tool learning~\cite{ToolLearning_qin2023tool}.  
There are some works primarily that concentrate on task-specific applications within specific domains~\cite{VisualChatGPT_wu2023visual,mm-react_yang2023mm,HuggingGPT_shen2023hugginggpt}.
Visual ChatGPT~\cite{VisualChatGPT_wu2023visual} seamlessly integrates ChatGPT with off-the-shelf visual models, leveraging ChatGPT as the agent and employing visual models as tools. This integration significantly expands its capabilities beyond text-based tasks, enabling Visual ChatGPT to handle tasks reliant on visual data effectively.
HuggingGPT~\cite{HuggingGPT_shen2023hugginggpt} employs ChatGPT to comprehend user instructions and planned tasks. Subsequently, it selects expert models hosted on Hugging Face to address the task and respond to the user.
Besides these works, some research focuses on how to use tools. 
VISPROG~\cite{VisProg_gupta2023visual} employs large language models to generate visual programs directly from instructions, bypassing the need for parsing language model outputs. This visual programming approach leverages LLMs' reasoning abilities for complex visual tasks, ensuring robust performance and producing interpretable visual rationales.
Toolformer~\cite{Toolformer_schick2023toolformer} learns how to use tools in a self-supervised way without requiring large amounts of human annotations.
ToolkenGPT~\cite{ToolkenGPT_hao2023toolkengpt} offers a novel solution, enabling large language models (LLMs) to proficiently handle diverse tools without requiring fine-tuning. It ensures rapid adaptation to new tools by representing each tool as a unique token, referred to as `toolken', expanding the model's vocabulary.

\paragraph{Prompting LLMs for Reasoning.}
Prompt engineering is an emerging discipline focused on the development and optimization of prompts to enhance the efficient utilization of large language models across a diverse array of applications and research domains.
Chain of Thought (CoT) prompting ~\cite{ChainOfThought_wei2022chain}, enhances large language models' reasoning by guiding them through step-by-step problem-solving, instead of directly providing answers. This method improves both transparency and accuracy in model responses.
In contrast to the original CoT technique, which requires the manual crafting of task-specific examples, the zero-shot Chain of Thought~\cite{StepByStep_kojima2022large} approach enables diverse reasoning tasks by simply appending ``Let's think step by step." to the prompt.
Based on zero-shot CoT, Auto CoT~\cite{AutoCoT_zhang2022automatic} generates reasoning chains to construct examples and then uses CoT reasoning method. 
ReAct ~\cite{ReACT_yao2022react} is a framework that leverages Large Language Models (LLMs) to generate reasoning traces and perform task-specific actions concurrently. This approach involves an iterative cycle where LLMs process observations, engage in decision-making and execute actions, which then lead to further observations and thoughts. Reasoning traces enable LLMs to plan and adapt actions, while their ability to interact with external tools enhances real-time response and contextual understanding
However, erroneous actions can lead to the propagation of mistakes, potentially trapping the model in a loop of continuous errors, such as repeatedly making incorrect API calls or generating incorrect responses.
For this reason, CoT and ReACT prompting techniques may fall short for complicated tasks.
To address this, some researchers are exploring the replacement of linear, chain-like reasoning methods with a tree-structured approach ~\cite{TreeOfThought_yao2023tree, toolllama_qin2023toolllm}.
Tree of Thought(ToT)~\cite{TreeOfThought_yao2023tree} promotes exploration by integrating intermediate thoughts, enabling comprehensive problem-solving with language models. 
Depth First Search-based Decision Tree (DFSDT)~\cite{toolllama_qin2023toolllm} is a specialized adaptation of the Tree of Thought framework, exclusively employing Depth First Search (DFS) to tackle complex decision-making issues in an unbounded decision space. However, DFSDT may ignore the answer information from another branch, which can lead to redundant actions. Thus, in this work, we propose Sum2Act to address this problem.

\paragraph{Large MultiModal Models.}
LLMs exhibit outstanding performance in handling textual data. However, their capabilities are inherently constrained when handling images. 
To address this limitation, many models~\cite{flamingo_alayrac2022flamingo,blip2_li2023blip2,LLaVA_liu2023visual,minigpt4_zhu2023minigpt,mplug-owl_ye2023mplug} focus on aligning image features with textual features. Combining these features through concatenation can boost LLMs' performance in various vision-language tasks, including visual question answering.
Seminal studies, such as VisualGPT~\cite{VisualGPT_chen2022visualgpt} and Frozen~\cite{Frozen_tsimpoukelli2021multimodal}, have highlighted the benefits of using pre-trained language models as vision-language decoders.
BLIP-2~\cite{blip2_li2023blip2} employs Q-Fromer to bridge the gap between image and text features.
Both LLaVA~\cite{LLaVA_liu2023visual} and MiniGPT-4~\cite{minigpt4_zhu2023minigpt} finetune the projection layer to better align extracted visual features from the frozen visual backbone with textual features.
mPLUG-Owl~\cite{mplug-owl_ye2023mplug} achieves superior performance in comprehending multi-modal instructions and participating in multi-turn dialogues through an innovative modularization-based training paradigm for large language models.
Large Multimodal Models are mostly used to understand and reason about data in different modalities, and also require a certain amount of data for training.
However, almost all kinds of these models are not training-free.

%% file: sec/3_method_new.tex
\section{Method}
This section presents our proposed Sum2Act, an efficient reasoning framework designed to enhance the ability of large language models to address real-world queries using extensive APIs. The overall architecture is first introduced in Sec.~\ref{sec-arch}. Then we detail two key components
in Sum2Act: action proposal module in
Sec.~\ref{sec-act} and summarization module in Sec.~\ref{sec-sum}.

\subsection{Overall Architecture}\label{sec-arch}
Sum2Act leverages a large language model and expansive open-world APIs to solve real-world tasks.
Sum2Act first utilizes a retriever to obtain relevant tools (or APIs) based on the correlation between users' instructions and descriptions or tools (or APIs).
Then it operates in two critical stages: the action proposal stage and the summarization stage, as depicted in Figure~\ref{fig:pipeline}.
Specifically, upon receiving a user's instruction, Sum2Act goes through an iterative cycle between the action proposal stage and the summarization stage. 
In the action proposal stage, the router evaluates its progress, determining whether to conclude the task or propose a tool for further action. 
Subsequently, in the summarization stage, the state manager assesses the observations of these actions, updating the overall state accordingly.
The process ends when Sum2Act completes the task and offers a solution. We briefly depict our algorithm in  \cref{alg:Sum2Act}.

\begin{algorithm}[th]
  \caption{Sum2Act}
  \label{alg:Sum2Act}
  \KwIn{User instruction: $I$.}
  \KwData{LLM: $\mathcal{M}$; API Retriever: $\mathcal{R}$; Initial state: $S_0$; Prompts: $\mathrm{prompt}$.}
  \KwOut{Answer: $Answer$.}
  \BlankLine
  Initialize state $S=S_0$\; 
  \tcp{0. Get relevant Tools}\
  $Tools=\mathcal{R}(I)$\\
  \While{True}{
  \tcp{1. Action Proposal Stage}
 $Action,Args=\mathcal{M}\left(\mathrm{prompt}_{\mathrm{plan}},I,S,Tools\right)$\;
  \If{$Action == ``\mathrm{Finish}"$}{
    {\bf break}\;
  }
  $observation=Action(Args)$;\\
  \tcp{2. Summarization Stage}
  $S=\mathcal{M}\left(\mathrm{prompt}_{\mathrm{state}}, I, S, observation,\right)$;\\
  }
  $Answer = Args[``\mathrm{Answer}"]$\;
\end{algorithm}

\begin{table*}
\resizebox{\linewidth}{!}{
    \centering
    \begin{tabular}{ccccccccccc}
        \toprule
        Model & Method & I1-Inst & I1-Tool & I1-Cat & I2-Inst & I2-Cat & I3-Inst & Average \\
        \midrule
        \multirow{3}*{ChatGPT} & ReAct-CoT~\cite{ReACT_yao2022react, ChainOfThought_wei2022chain} &36.0&52.0&40.0&42.5&39.0&37.0&41.1\\
        & DFSDT~\cite{toolllama_qin2023toolllm}&57.0&63.0&63.0&\bf78.0&\bf69.0&72.0& 67.0\\
        & Sum2Act(Ours)&\bf71.0&\bf71.0&\bf65.0&\bf78.0&61.0&\bf74.0&\bf70.0\\
      \bottomrule
      \end{tabular}}
       \caption{{\bf Pass Rate} on ToolBench. All methods use the Oracle API retriever(i.e., ground-truth APIs). Due to the APIs' dynamic nature, the results for ReAct and DFSDT with ChatGPT are obtained by rerunning their codes.}
       \label{tab:pass}
\end{table*}

\begin{table*}
\resizebox{\linewidth}{!}{
    \centering
    \begin{tabular}{ccccccccccc}
        \toprule
        Model & Method & I1-Inst & I1-Tool & I1-Cat & I2-Inst & I2-Cat & I3-Inst & Average \\
        \midrule
        \multirow{3}*{ChatGPT}
        & DFSDT~\cite{toolllama_qin2023toolllm} against ReAct-CoT~\cite{ReACT_yao2022react, ChainOfThought_wei2022chain}    & 63.5 & 54.5 & 65.0 & 70.0 & 69.0 & 72.5 & 65.8 \\
        & Sum2Act against ReAct-CoT~\cite{ReACT_yao2022react, ChainOfThought_wei2022chain} & 71.5 & 59.5 & 66.5 & 73.5 & 61.5 & 74.5 &\bf 67.8 \\
        & Sum2Act against DFSDT~\cite{toolllama_qin2023toolllm}     & 60.0 & 58.5 & 56.0 & 55.0 & 48.0 & 50.0 &\bf 54.6 \\
      \bottomrule
      \end{tabular}}
       \caption{{\bf Win Rate} on ToolBench. It is obtained by pairwise comparison among these methods. All methods use the Oracle API retriever(i.e., ground-truth APIs). Due to the APIs' dynamic nature, the results for ReAct and DFSDT with ChatGPT are obtained by rerunning their codes. Following ToolLLM~\cite{toolllama_qin2023toolllm}, we split the ratio of tie into two halves and add them to the win and lose rate. }
       \label{tab:compare}
\end{table*}

\begin{table*}
\resizebox{\linewidth}{!}{
    \centering
    \begin{tabular}{ccccccccccccccccc}
        \toprule
        \multirow{2}*{Model}& \multirow{2}*{Method}&\multicolumn{2}{c}{I1-Inst}&\multicolumn{2}{c}{I1-Tool}&\multicolumn{2}{c}{I1-Cat}&\multicolumn{2}{c}{I2-Inst}&\multicolumn{2}{c}{I2-Cat}&\multicolumn{2}{c}{I3-Inst}&\multicolumn{2}{c}{Average}\\
        \cmidrule(r){3-4}\cmidrule(r){5-6}\cmidrule(r){7-8}\cmidrule(r){9-10}\cmidrule(r){11-12}\cmidrule(r){13-14}\cmidrule(r){15-16}
        &&Pass & Win&Pass & Win&Pass & Win&Pass & Win&Pass & Win&Pass & Win&Pass & Win\\
        \midrule
        \multirow{3}*{ChatGPT} & Sum2Act w/o task decomp&\bf71.0&\bf71.5&71.0&59.5&65.0&66.5&\bf78.0&\bf73.5&61.0&61.5&\bf74.0&\bf74.5&70.0&67.8\\
        & Sum2Act+task decomp. &62.0&64.0&\bf75.0&\bf61.0&\bf73.0&\bf74.5&73.0&70.5&\bf67.0&\bf68.5&\bf74.0&74.0&\bf70.7&\bf68.8\\
      \bottomrule
      \end{tabular}}
       \caption{Ablation study to check the influence of task decomposition on tool invocation. When task decomposition is adopted, the LLM will first decompose the target task into several subtasks according to the available tools. Then, the decomposed tasks are attached to the prompts as guidance.}
       \label{tab:ablation}
\end{table*}

\subsection{Action Proposal}\label{sec-act}
At the action proposal stage, the router will plan the next action and then execute it. 
Specifically, we employ a language model $\mathcal{M}$ and well-designed prompt $\mathrm{prompt}_{\mathrm{plan}}$ to act as the router. 
The router engages in thought and reasoning based on the current state, the instruction, and available tools, subsequently choosing the next action and the arguments of the action.
This process can be formalized as follows:
\begin{equation}
{Action},{Args} = \mathcal{M}(\mathrm{prompt}_{\mathrm{plan}},{I},{S},Tools)
\end{equation}
$Action$ can represent either one of the tools or the special action $``\mathrm{Finish}"$, signifying the completion of the task.
\begin{itemize}
    \item If the value of $Action$ is not ``$\mathrm{Finish}$", it indicates that, from the router's perspective, the task has not been completed. In this scenario, $Action$ denotes a specific tool or API, and $Args$ encapsulates the corresponding arguments for the given $Action$. The router proceeds to execute the specified action, subsequently obtaining relevant observations in the process.
    \begin{equation}
        observation = Action\left(Args\right)
    \end{equation}
    \item If the value of $Action$ is exactly ``$\mathrm{Finish}$", it indicates that the router perceives the task as completed. In such a scenario, the program will exit the loop, respond to the user's command, and ultimately terminate. The variable $Args$ contains the answer utilized to respond to the user. 
    \begin{equation}
        Answer = Args[``\mathrm{Answer}"]
    \end{equation}
\end{itemize}

\subsection{Summarization}\label{sec-sum}
After the action proposal stage, Sum2Act transitions into a novel stage we have termed the ``summarization" stage. 
At this stage, the central component is the state manager, which is responsible for maintaining the {\bf State} of the system.

We draw inspiration from the human working process: when an individual is engaged in a complex task, they periodically summarize their progress, noting what has been completed and challenges encountered. The summarization process involves effectively processing the outcomes of past actions, highlighting significant results, filtering out task-irrelevant information, and aiding in subsequent planning for the individual.

Distinct from the conventional memory structure used in many reasoning methods, the {\bf State} in our system is a dynamic representation of the current scenario, informed by all observations of preceding actions.
This differs from traditional memory, which primarily logs observations from past actions without an evolving contextual understanding.
In each step, our state manager will check whether the new action successfully returns information related to the target task. If the new action succeeds, the state manager will register the newly obtained answer, otherwise, it will figure out the failed reason and add the failed action to the failure history. 

Therefore, the task {\bf State} comprises two primary elements:
\begin{itemize}
\item {\it Current results}: This aspect reflects the ongoing progress toward task completion.
\item {\it Failure history}: This records any issues encountered with tools, guiding the router to avoid repeat selections of problematic options.
\end{itemize}

In the Chain of Thought (CoT) process, the length of memory can become cumbersome as observations accumulate. Practical applications often resort to truncation methods to manage this, but such approaches invariably lead to information loss. This presents a fundamental trade-off: balancing the completeness of memory with its manageability.

The excessively elongated memory can impede efficiency and decision-making ability of the router due to the overload of irrelevant data. 
On the other hand, overly truncated memory can leave the router under-informed, especially when dealing with complex tasks.

To address this, the state manager synthesizes the observation and updates the {\bf State} accordingly after each action executed by the router. The process is formalized as follows:
\begin{equation}
S = \mathcal{M}(\mathrm{prompt}_{\mathrm{state}}, I, S, {observation})
\end{equation}
Here, ${observation}$ denotes the outcome of the tool's execution, while $\mathrm{prompt}_{\mathrm{state}}$ is the associated prompt. 
Contrary to the memory, the length of textual content within a state is maintained within reasonable limits. It encapsulates a summary of all previous actions' observations, thus minimizing information loss. Consequently, the information density within the state is typically much higher than that in memory, a crucial feature in open-world scenarios where APIs often yield extensive but irrelevant data.

Upon summarizing the observations and updating the {\bf State}, the state manager efficiently streamlines the decision-making process for subsequent actions.

%% file: sec/4_exp.tex
\section{Experiments}
In this section, experiments are conducted to evaluate the effectiveness of our proposed Sum2Act. 
In Sec.~\ref{sec-toolbench}, we compare our method with existing baselines on ToolBench~\cite{toolllama_qin2023toolllm} datasets, which contains more than 16000 real-time APIs, to demonstrate the capability of Sum2Act in addressing complicate tasks. Furthermore, we incorporate additional visual APIs with ToolBench in Sec.~\ref{sec-visual}, revealing its ability to handle more diverse vision tasks.

\subsection{Sum2Act for Open-World task solution}\label{sec-toolbench}

\begin{figure*}[thb]
\begin{center}
   \includegraphics[width=1.0\linewidth]{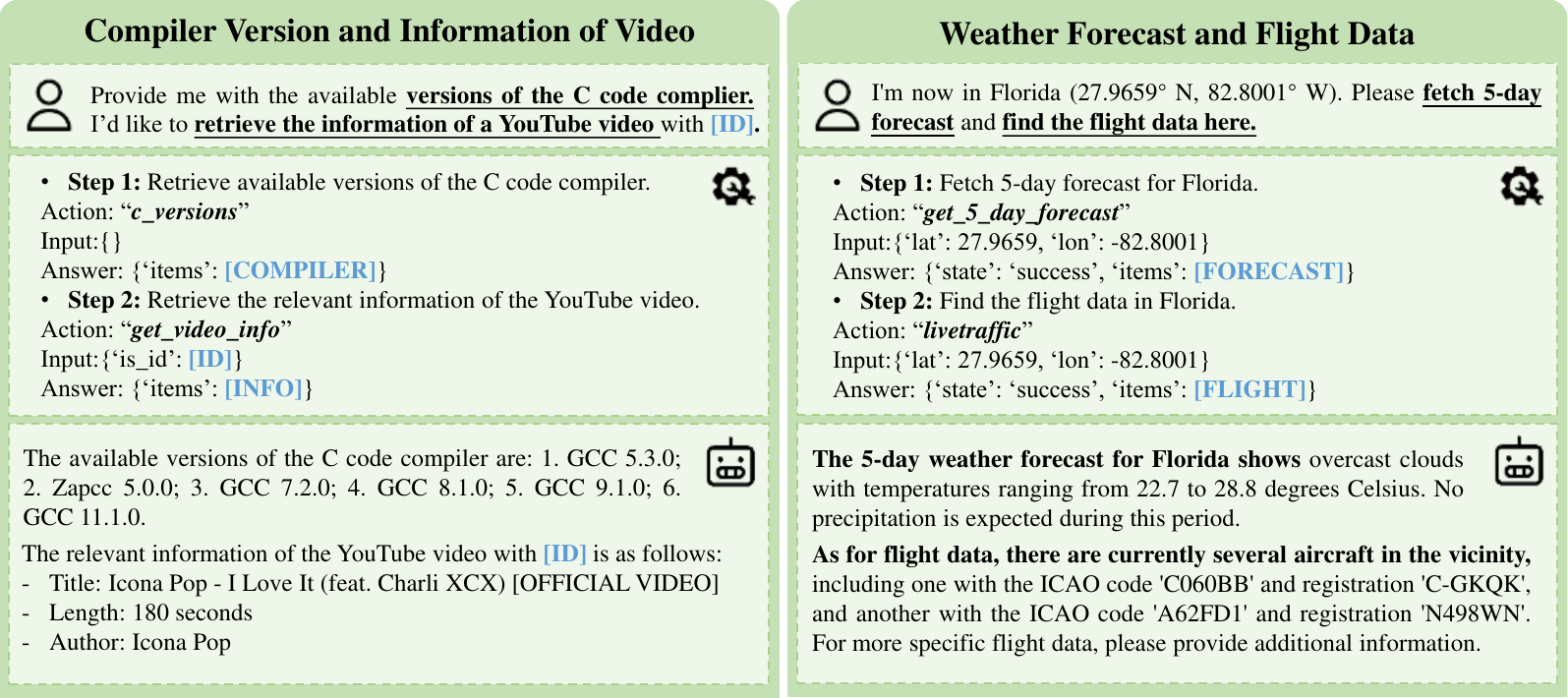}
\end{center}
   \vspace{-2mm}
   \caption{Case Studies: Utilizing Open-World APIs for Diverse Queries. The left scenario demonstrates querying for C compiler versions and video information, while the right scenario showcases the use of specific APIs to gather Florida's weather forecast and flight data.}
   \label{fig:demo_vis_1}
   \vspace{-2mm}
\end{figure*}

\paragraph{Datasets}
We evaluate Sum2Act's Open-world task solution performance on ToolBench~\cite{toolllama_qin2023toolllm} Datasets.
This dataset collected more than 16000 real-time APIs from RapidAPI Hub, which are from 49 coarse-grained categories and 500+ more fine-grained categories called \textbf{collections}.
The test set consists of instructions spanning three distinct levels of difficulty:
(1) {\bf Inst.}: unseen instructions from the same set of tools in the training data (We don't use the training data in this paper),
(2) {\bf Tool.}: instructions that require unseen tools belonging to the same category,
(3) {\bf Cat.}: instructions that require unseen tools belonging to different categories.

The evaluation is conducted regarding 3 scenarios: single-tool instructions (I1), intra-category multi-tool instructions (I2),
and intra-collection multi-tool instructions (I3).
The final evaluation protocol results in 6 subsets for testing, which are I1-Inst, I1-Tool, I1-Cat, I2-Inst, I2-Cat, and I3-Inst respectively.
Each of these six subsets contains 100 test examples.

\paragraph{Evaluation metrics}
ToolEval~\cite{toolllama_qin2023toolllm} proposes two evaluation metrics, which are {\bf Pass Rate} and {\bf Win Rate} respectively.
{\bf Pass Rate} reflects the proportion of successfully completing an instruction within a limited number of actions( The maximum step numbers set in literature \cite{toolllama_qin2023toolllm} is 200 for DFSDT, while in this paper we require Sum2Act to finish within 30 steps). 
The metric measures whether the obtained answer is relevant to the user's instruction. However, it does not measure how well it is completed.
{\bf Win Rate} takes the total execution steps, answer quality, and the diversity of used APIs into consideration to evaluate how well a task is accomplished. It is measured by comparing two solution paths for a given instruction using an evaluator. Following ToolLLM~\cite{toolllama_qin2023toolllm}, we calculate the Win Rate using ChatGPT as the evaluator.

\paragraph{Settings} To eliminate the influence posed by API retriever, we directly use the ground-truth (oracle) APIs provided in ToolBench. Thus, the task-solving performance only depends on the prompting methods and the LLM itself. We test our method with ChatGPT.

\begin{figure*}[t]
\begin{center}
   \includegraphics[width=1.0\linewidth]{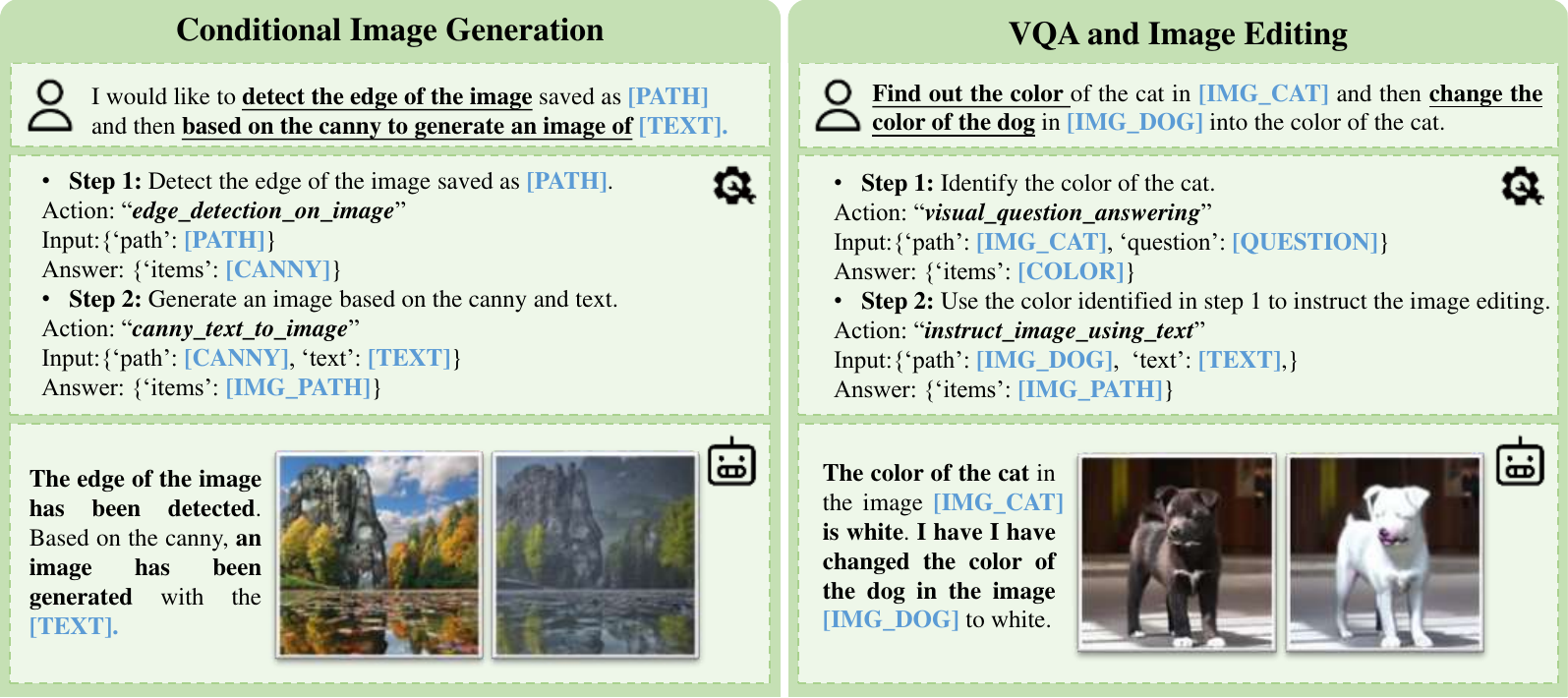}
\end{center}
    \vspace{-2mm}
   \caption{Case Studies: Demonstrating Sum2Act's versatility in vision tasks, showcasing its proficiency in conditional image generation and its ability to integrate multiple vision APIs for targeted image editing.}
   \label{fig:demo_vis_2}
   \vspace{-2mm}
\end{figure*}

\begin{figure*}[t]
\begin{center}
   \includegraphics[width=1.0\linewidth]{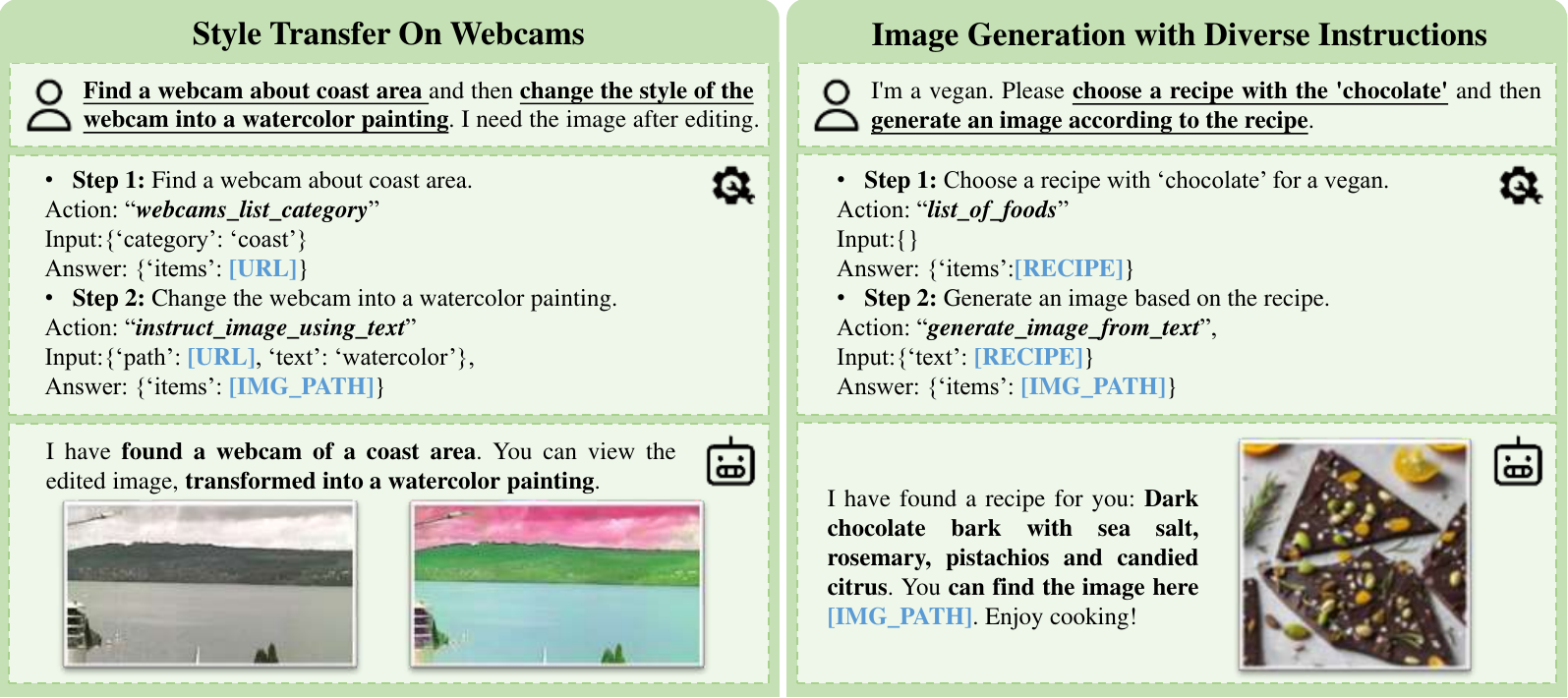}
\end{center}
    \vspace{-2mm}
   \caption{Case Studies: Illustrating Sum2Act's adeptness in handling intricate tasks through a combination of open-world and vision APIs. On the left, a webcam image undergoes style transfer, while on the right, an image is generated that corresponds to a searched vegan recipe. This highlights Sum2Act's proficiency in seamlessly integrating both open-world and varied vision APIs.}
   \label{fig:demo_vis_3}
\end{figure*}

\paragraph{Main Results}
In~\Cref{tab:pass} and~\Cref{tab:compare}, we present a comparative analysis of three distinct methods evaluated across six test sets from ToolBench, labeled as I1-Inst, I1-Tool, etc.
As for {\bf Pass Rate}, the ReACT-CoT method gets the lowest average rate in our study, standing at 41.1\%.
In contrast, the DFSDT method exhibits consistent performance across almost all test scenarios, achieving an average rate of 67.0\%.
Our proposed method, Sum2Act, surpasses both ReACT-CoT and DFSDT, recording an impressive pass rate of 70.0\%.
When considering the {\bf Win Rate} metric, our method achieves rates of 67.8\% against ReAct-CoT and 54.6\% against DFSDT. Notably, both outcomes exceed the 50\% threshold, signifying our superior performance.
This superior performance underscores Sum2Act's effectiveness and adaptability, reinforcing its standing as the most proficient method in our analysis.

The overall analysis presents a clear hierarchy of method performance. Sum2Act emerges as the most robust and effective pipeline, showcasing high efficiency and adaptability across diverse testing scenarios.
DFSDT, while consistent and reliable, falls short of the benchmark set by Sum2Act. 
On the other end of the spectrum, ReAct significantly underperforms when compared to both DFSDT and Sum2Act.
This underperformance is particularly noteworthy as it underscores the inherent limitations of simpler, chain-based reasoning methods in tackling complex, multi-faceted tasks.
The findings from our study emphasize the critical need for models that incorporate a broader search space and possess advanced error-handling capabilities
These attributes are essential for efficient and effective tool invocation, especially in scenarios that demand a higher degree of complexity and variability.

In the referenced cases illustrated in~\Cref{fig:demo_vis_2}, the effectiveness of Sum2Act is prominently displayed. Each example demonstrates the router's capability to accurately interpret user instructions and execute appropriate actions. 
The cases highlight instances where the router not only identifies the correct course of action based on the given instructions but also effectively communicates its responses back to the user. 
This successful interaction underscores the proficiency of Sum2Act in understanding and acting upon complex user directives, showcasing the practical application of our framework in real-world scenarios.

\paragraph{Should we decompose the target task first?}
Recent literature~\cite{ControlLLM_2023controlllm} has adopted task decomposition before starting to call APIs. In this work, we also try to verify whether a task decomposition module can help LLM to better solve complex tasks. We instruct ChatGPT to break the target task into several subtasks according to the available tools and then attach the decomposed tasks to the prompts of the router as guidance. The final results are reported in \cref{tab:ablation}. As shown in the table, a task decomposition module can slightly improve the average Pass Rate and Win Rate, however, this improvement is not significant. This is reasonable since accurate task decomposition itself is still an open problem. Further, how to guide the task-solving process based on the decomposed tasks is also an interesting research direction.

\subsection{Integrating Sum2Act with Visual APIs.}\label{sec-visual}

The Toolbench benchmark comprises a total of 49 categories of APIs, with the majority oriented towards text-based and query-related functionalities.
Expanding upon this existing framework, we have incorporated supplementary APIs of visual processing, facilitating advanced operations specifically designed for image manipulation.
SDXL~\cite{sdxl_podell2023sdxl} model is employed for image generation, ControlNet for conditional generation, Blip~\cite{blip_li2022blip} for Visual Question Answering (VQA), InstructPix2Pix~\cite{instructpix2pix_brooks2023instructpix2pix} for image editing and \etc.

Through the integration of visual APIs, our model acquires heightened capabilities, facilitating the execution of intricate tasks within the realm of image manipulation.
In~\Cref{fig:demo_vis_2}, the model exclusively employs visual APIs to perform diverse image-related tasks. In the first case, it undertakes conditional image generation, while in the second case, it utilizes the Visual Question Answering (VQA) tool to inquire about the color of a cat and subsequently alters the color of a dog to match that of the cat.

As depicted in~\Cref{fig:demo_vis_3}, our model can adeptly integrate APIs sourced from Toolbench alongside visual APIs. In the first scenario, it can find a webcam and then perform the style transfer. In the second scenario, it queries vegan recipes, selects one meeting a specified condition (e.g., 'with chocolate'), and subsequently generates a high-quality image corresponding to the chosen recipe.

%% file: sec/5_conclusion.tex
\section{Discussion}

Sum2Act represents an efficient reasoning framework that synergizes the simplicity of methods like CoT or ReAct with the search capabilities of tree-based reasoning approaches such as ToT and DFSDT.
Sum2Act operates like ReAct, proposing thoughts and actions based on the current progress of the task.
The distinctive feature of Sum2Act is its state manager, capable of validating the outcomes of actions and pinpointing the reasons behind failures.
This feature ensures that Sum2Act consistently maintains an accurate understanding of the current task states, effectively avoiding the error propagation commonly observed in CoT and ReAct. 
Moreover, Sum2Act stands out from CoT and ReAct by not being limited to a linear path. It can explore multiple directions to find needed information and ensure that unsuccessful actions don't negatively impact the router's decision-making. 
This multi-directional approach gives Sum2Act a wide search range similar to tree-based methods. 
However, unlike these methods, which can miss information from previous paths and repeat actions, Sum2Act avoids this redundancy.
Sum2Act, with its balanced approach, offers a more efficient and comprehensive search strategy.


\section{Conclusion}


In our research, we present Sum2Act, an innovative reasoning framework developed to augment the capabilities of LLMs in tackling real-world queries. Sum2Act stands out for its ability to seamlessly integrate massive open-world APIs with vision APIs, allowing LLMs to process visual data alongside textual data. 
The essence of Sum2Act lies in its process, which maintains a continual awareness within LLMs about the target task and its progression.
Central to our approach is the requirement for LLMs to summarize historical information. 
This involves understanding which subtasks have been completed, identifying reasons for any failures, and using this summary to inform subsequent steps. 
Despite its apparent simplicity, Sum2Act has exhibited remarkable performance, outshining established methods like ReAct and DFSDT in ToolBench evaluations.

%% file: sec/X_suppl.tex
\maketitlesupplementary

\section{More Details of Prompts}
This section provides detailed information about the prompts we used in Sum2Act, with a specific focus on the roles of the Router and State Manager. These prompts are crafted to assist the Router in making decisions and to enable the State Manager to manage the system's state efficiently.

\begin{figure*}[hbt]
\begin{center}
   \includegraphics[width=0.95\linewidth]{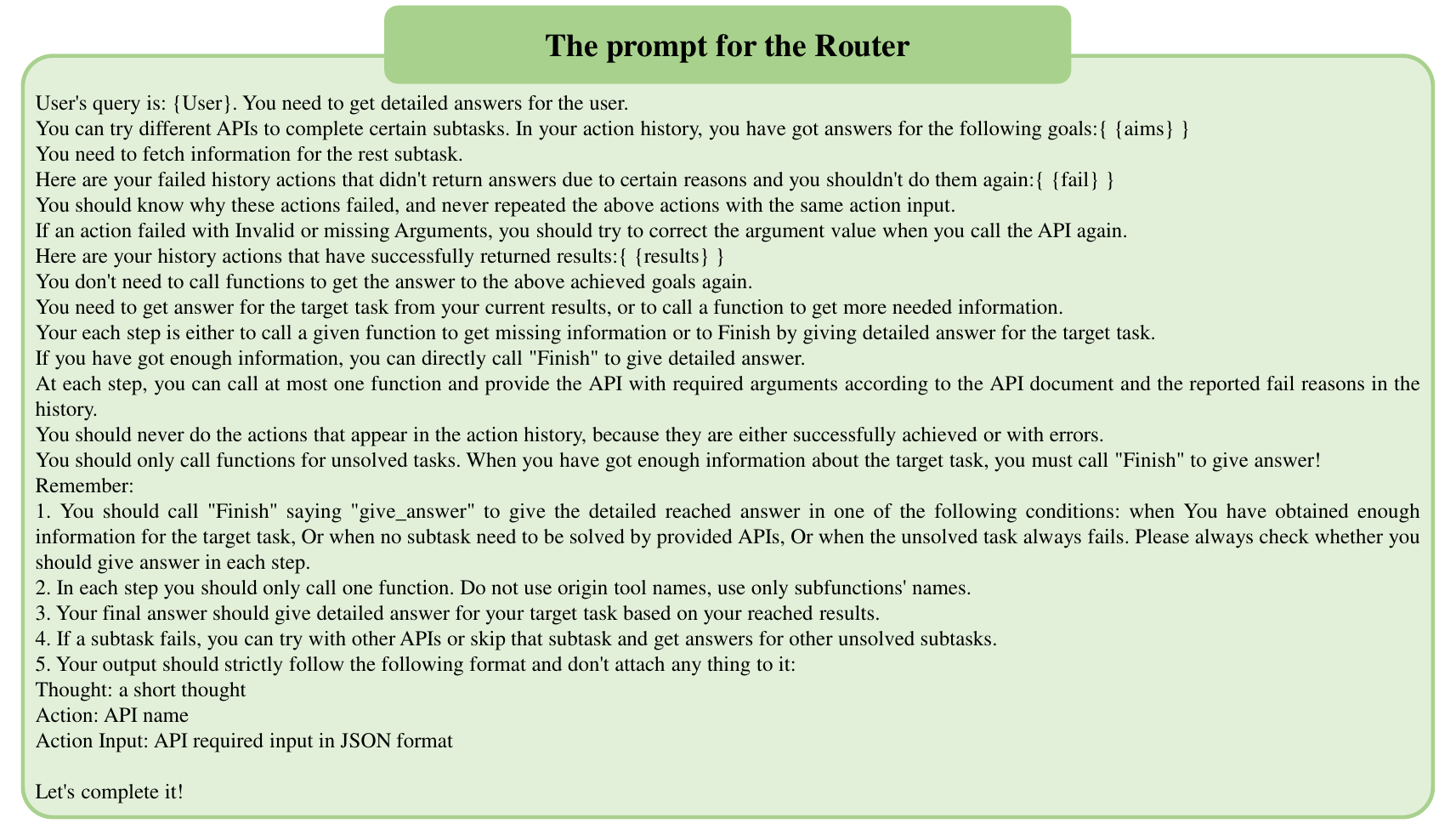}
\end{center}
\label{fig:router_prompt}
\caption{The prompt for the Router}
\end{figure*}

\subsection{Prompt for the Router}
The prompt structure for the Router can be divided into three distinct parts, each serving a unique purpose in the process:

\begin{itemize}
\item \textbf{User Instruction}: This is the initial query or command provided by the user. It forms the basis of the task and is essential for defining the Router's objective.
\item \textbf{State}: The State component is a dynamic element of the prompt, encompassing two critical aspects:
    \begin{enumerate}
        \item \textit{Current Results}: This part reflects the ongoing progress of the task. It includes information about the subtasks that have been completed, the results obtained so far, and any partial achievements. The current results help Router access how much of the task has been accomplished and what remains to be done.
        \item \textit{Failure History}: An account of the unsuccessful attempts or errors encountered in previous steps. This historical data is invaluable for Router to avoid repeating past mistakes or selecting tools that have previously proven ineffective.
    \end{enumerate}

\item \textbf{Others}: This section of the prompt includes additional rules that the Router should adhere to during the action proposal stage. These might involve constraints, preferences, or specific methodologies to be followed. Also included is the desired format for the Router's output, which dictates how the results should be presented or communicated.
\end{itemize}
Every component of the prompt is essential for directing the Router's decision-making process, guaranteeing a systematic and effective method for suggesting a new action. The prompt is illustrated in~\Cref{fig:router_prompt}.

\subsection{Prompt for the State Manager}
\begin{figure*}[t]
\begin{center}
   \includegraphics[width=0.95\linewidth]{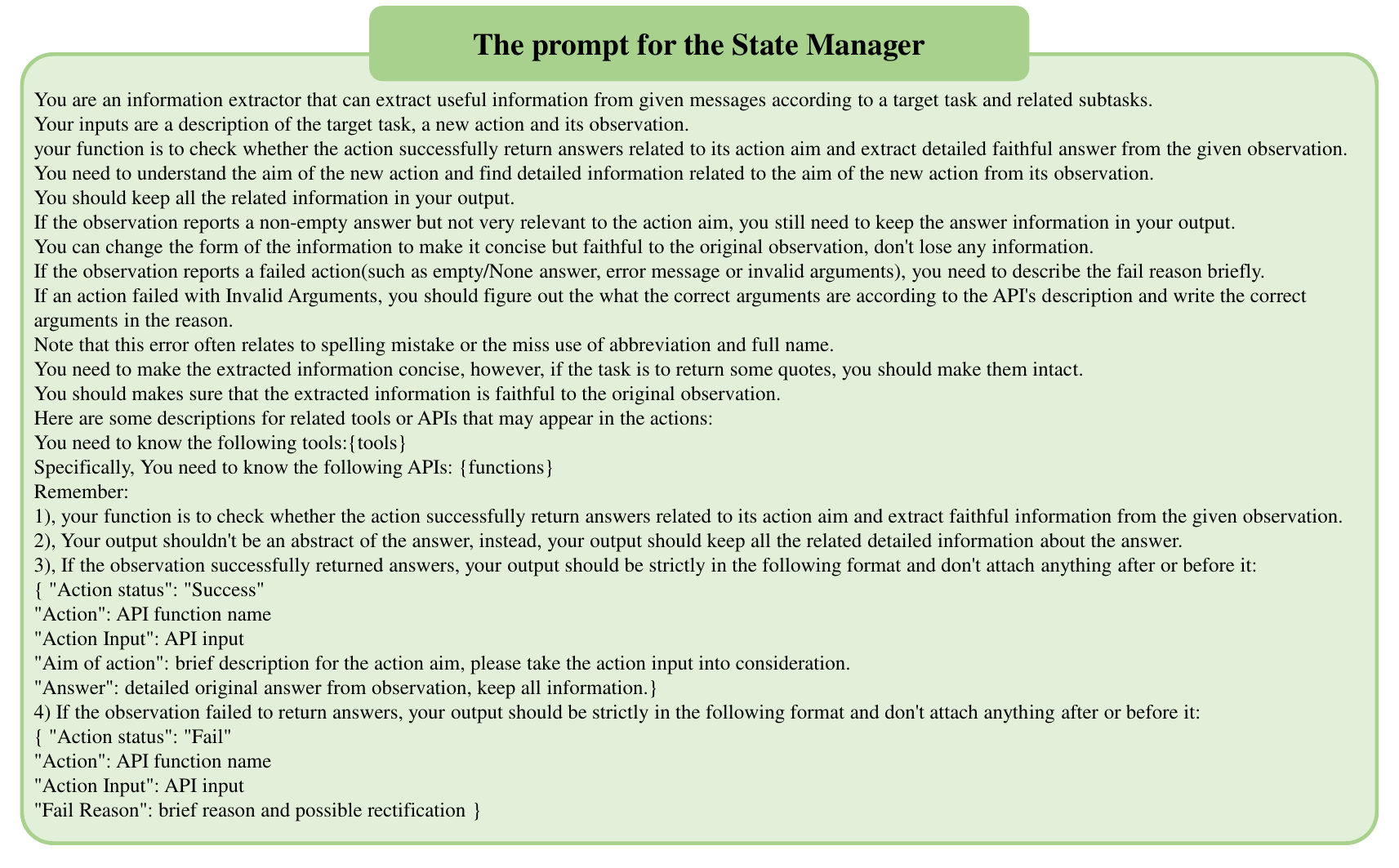}
\end{center}
\label{fig:state_prompt}
\caption{The prompt for the State Manager.}
\end{figure*}

The prompts for the State Manager also comprise three distinct parts, each tailored to facilitate effective state management:

\begin{itemize}
\item \textbf{Summarization}: This involves distilling key observations from successful tool usage. The State Manager should filter out irrelevant or redundant information to maintain a clear and concise record of successful outcomes and insights.
\item \textbf{Dealing with Failure}: In instances of failure, this part of the prompt guides the State Manager to analyze and deduce possible reasons for the failure. Understanding the causes of unsuccessful actions is crucial for refining future strategies and avoiding similar pitfalls.
\item \textbf{Others}: Similar to the Router, this section includes operational rules and guidelines specific to the State Manager. It also specifies the desired format for the State Manager's outputs, ensuring clarity and consistency in how the state information is updated and presented.
\end{itemize}
Each component of these prompts plays a critical role for the State Manager.
They help the State Manager better maintain the state. The prompt can be viewed in~\Cref{fig:state_prompt}.

Both prompts are significant for the overall performance of Sum2Act, ensuring that both the Router and the State Manager operate efficiently and effectively within their designated roles. 

